\pdfoutput=1
\documentclass[11pt]{article}

\usepackage{emnlp2021}

\usepackage{times}
\usepackage{latexsym}

\usepackage[T1]{fontenc}
\usepackage{microtype}

\usepackage{graphicx}
\usepackage{cleveref}
\usepackage{array}
\usepackage{tabularx}
\usepackage{multirow}

\title{Domain Adaptation of a State of the Art Text-to-SQL Model: Lessons Learned and Challenges Found}

\author{Irene Manotas \\
  IBM Research\\
  \texttt{irene.Manotas@ibm.com} 
\And
  Octavian Popescu \\
IBM Research \\
  \texttt{o.popescu@us.ibm.com} 
\AND
  Ngoc Phuoc An Vo\\
  IBM Research\\
  \texttt{ngoc.phuoc.an.vo@ibm.com}
\And
  Vadim Sheinin\\
  IBM Research\\
  \texttt{vadims@us.ibm.com}}

\begin{document}
\maketitle

\begin{abstract}

There are many recent advanced developments for the Text-to-SQL task, where the Picard model is one of the the top performing models as measured by the Spider dataset competition. However, bringing Text-to-SQL systems to realistic use-cases through domain adaptation remains a tough challenge. We analyze how well the base T5 Language Model and Picard perform on query structures different from the Spider dataset, we fine-tuned the base model on the Spider data and on independent databases (DB). To avoid accessing the DB content online during inference, we also present an alternative way to disambiguate the values in an input question using a rule-based approach that relies on an intermediate representation of the semantic concepts of an input question. In our results we show in what cases T5 and Picard can deliver good performance, we share the lessons learned, and discuss current domain adaptation challenges.

\end{abstract}

\section{Introduction}
\label{intro}

Semantic parsing is an important NLP task where natural
language utterances are mapped to formal meaning representations. Text-to-SQL datasets and models have been built to support the semantic parsing for SQL, and are crucial to support building natural language interfaces to databases.
Transfer Learning (TL), a technique where a pre-trained model on a data-rich task is then fine-tuned on a downstream task, has emerged as a powerful mechanism to train models for different NLP tasks, including Text-to-SQL. Leveraging the transformer architecture \citep{Vaswani2017:Transformer} several models have been able to achieve state of the art performance for the semantic parsing task, such as the BERT model \citep{Devlin2019:BERT}.
Inspired by the success of pre-trained models on a data-rich task, \citeauthor{Raffel2020:T5} presented a unified approach to systematically study different TL approaches, treating every NLP problem as a "text-to-text" problem, and introducing the T5 model \citep{Raffel2020:T5}.
The Parsing Incrementally for Constrained Auto-Regressive Decoding (Picard) model, proposed by \citeauthor{Scholak2021:PICARD}, enhances the output of pre-trained language models for the text-to-SQL task, like T5, by finding valid output sequences, and rejecting invalid tokens at each decoding step via incremental parsing. Picard is so far one of the top models on the Spider dataset, the largest cross-domain dataset for Tex-to-SQL.

Domain generalization (a.k.a cross-domain generalization) is a realistic and important challenge in the Text-to-SQL task. It addresses whether a model being trained on a dataset and accompanied DBs, still can perform comparably well on questions from other independent dataset and DB. Especially it is practical in real scenarios whenever we want to use a Text-to-SQL model for a new DB without sufficient questions and SQL queries for training because of the cost associated with the annotation of new questions and SQL queries for new DBs.
Although both T5 and Picard achieved remarkable performances on Spider, to the best of our knowledge, there has not been an attempt to evaluate these SOTA models on independent datasets and DBs outside of Spider.

In this paper we evaluated how the pre-trained T5 model and the Picard model perform on independent DB that are different in terms of questions, queries, and schema, to those included in the popular Spider dataset \citep{Yu2018:spider}.
We empirically investigated and analyzed both the T5 and Picard models by evaluating them in a zero-shot setting and by fine-tuning the pre-trained Spider models on a set of independent schemas. 

Our results indicate that for queries having certain characteristics, T5 with Picard give the best performance, but there are still challenges with complex and realistic queries where using the T5 model without Picard performs better, e.g., with queries using SQL functions or with multiple conditions. We present the lessons learned while fine-tuning the models to a set of independent DB, and the open challenges related to queries having certain features, as well as with the evaluation metrics.

\section{Related Work}
\label{sec:relwork}

Text-to-SQL is a long standing and an important NLP task with many references on its complexity and achievements obtained recently \cite{Pop2003:NLIDB, Ziyu2010, Navid2017}. 
The Spider dataset \cite{Yu2018:spider} was introduced a few years ago as a large-scale complex and cross-domain semantic parsing and Text-to-SQL annotated dataset. This dataset, and its challenge, has motivated the development and evaluation of text-to-SQL models. Teams participating in the Spider challenge can train their models on the Spider training data and then evaluate them on an unseen test set, not available to the public. Spider train set has \(7,000\) example questions and queries for \(200\) databases covering \(138\) different domains. DBs in the Spider dataset can have multiple tables and complex queries.

 \citeauthor{Xie2022:UnifiedSKG} designed a framework to benchmark Structured knowledge grounding (SKG) for different tasks, including text-to-SQL. Using as a base the T5 model the authors showed that T5 achieves state-of-the-art performance on almost all of the studied \(21\) tasks. \citep{Xie2022:UnifiedSKG}. Picard \cite{Scholak2021:PICARD}, leveraged LMs and proposed a mechanism to constraining auto-regressive decoders via incremental parsing, which using T5 as a base model, put Picard at the top of the rank in the Spider challenge.

\paragraph{Domain Adaptation.} To test the domain adaptation capability of a Text-to-SQL model, the model is trained on examples of training DBs and evaluated on examples of the evaluation DBs that are not seen during training phase \cite{yu2019cosql,yu2019sparc}. Several studies \cite{suhr2020exploring, deng2020structure,gan2021towards} show that domain adaptation remains a big challenge due to generalization performance of Text-to-SQL models, particularly the models trained on a cross-domain dataset such as Spider \cite{Yu2018:spider}
do not generalize to a new external database. The study \cite{suhr2020exploring} presented some unexplored generalization challenges in domain adaptation for the Text-to-SQL task due to the distribution shift of both questions and SQL queries. Another study \cite{gan2021exploring} showed that the generalization performance could be low even for both questions and SQL queries having similar distribution to the training set.

\section{DBs and Special SQL Queries}
\label{sec:queries}

In this section we present the Databases and SQL queries we used to analyze the performance of the T5 and Picard models.

\begin{table}[h!]
\centering
\begin{tabular}{||c | c | c | c||}
\hline
DB & Train & Dev & Test \\ 
\hline\hline
HR & 99  & 10 & 78 \\ %& \\
\hline
WH & 146 & 16 & 40 \\ %& \\
\hline
IN & 145 & 18 & 46 \\ %& \\
\hline
\end{tabular}
\caption{Data splits for independent schemas having questions and query examples in SQLite.}
\label{tab:splits}
\end{table}

\subsection{Independent DBs}
\label{sec:dbs}

We used three independent DBs outside of the Spider dataset for our analysis:
\paragraph{Human Resource (HR).}  A one-table database containing employees information such as employee number, name, birthdate, hire date, leave date, department, manager, salary, bonus.

\paragraph{Warehouse (WH).} A complex schema for sales activities consisting of eight tables including vendors, shops, stock, customers, manufacturers, products, sales, and sales details.

\paragraph{Invoicing (IN).} A complex schema for recording invoicing tasks that comprises four tables including record, contract, project, and assignment.

Table \ref{tab:splits} shows the data splits for each of the three different DBs we considered. The examples in each of the splits have queries with a similar structure to those in the Spider dataset, both simple and complex queries, as well as some new SQL structures such as multiple disjoint conditions and non-explicit column mentions for all questions. To reduce possible false positives for the EA evaluation metric, we verified that all queries returned non-empty values. 

\begin{figure}[t]
    \centering
         \includegraphics[width=1.3\columnwidth]{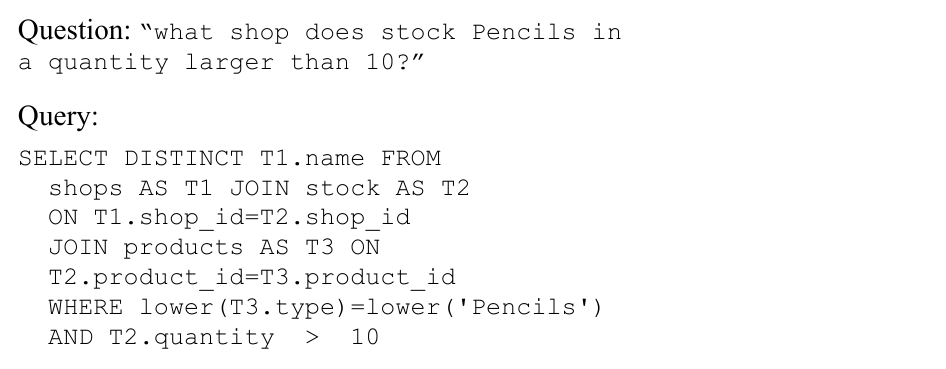}
         \caption{Examples of SQL query using the String function \textit{lower()}}
         \label{fig:queryLower}
\end{figure}

\begin{figure}[t]
        \centering
         \includegraphics[width=0.9\columnwidth]{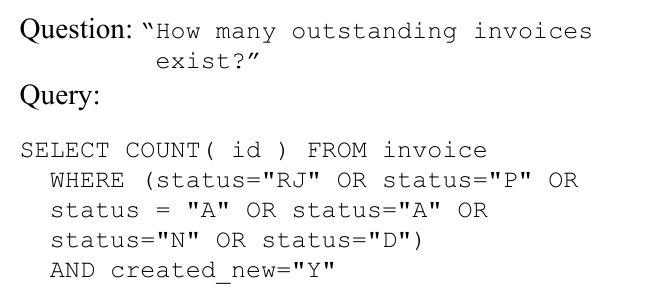}
         \caption{Examples of SQL query having multiple SQL OR conditions.}
         \label{fig:queryOR}
\end{figure}

\begin{figure}[t]
        \centering
         \includegraphics[width=1.1\columnwidth]{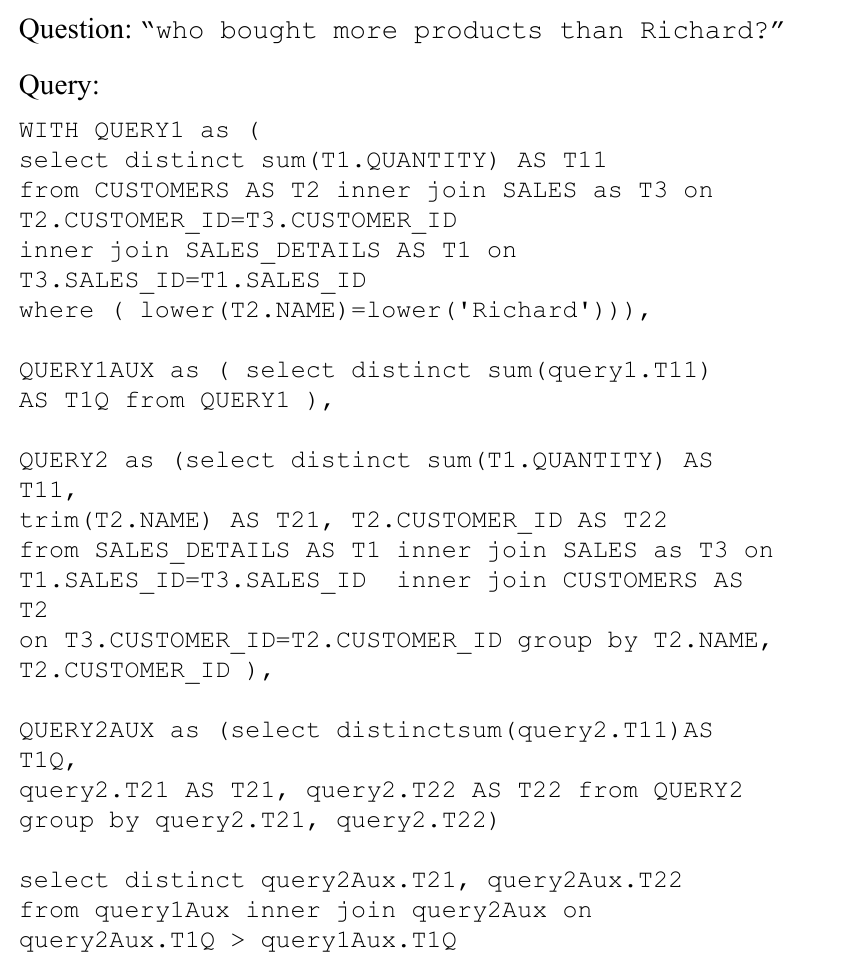}
         \caption{Example of SQL CTE query from the WH DB.}
         \label{fig:with_query}
\end{figure}

In addition to analyzing how well the selected models performed in general questions and SQL queries for these three independent DBs, we  were also interested in investigating how string functions, sub-query refactoring, and multiple disjoint conditions, were learned by these models. We refer to these types of queries as \textit{special} queries. The types of SQL language elements in the set of \textit{special} queries are commonly used in practice and therefore are important in a pragmatic setting.
We analyzed whether or not \textit{special} queries were considered in the Spider dataset and found that there were no examples of these queries in the train or development sets.

\subsection{Queries With SQL String Functions}

Using functions in SQL is a common practice.
String functions in SQL are used to extract, transform, or to return information about a string. 
To evaluate how the T5 and Picard models could learn and predict String functions in queries, we selected two string functions and transformed the queries in the WH DB to include the functions \textit{trim()} and \textit{lower()} when possible. \Cref{fig:queryLower} shows an example query from the WH DB using the \textit{lower} function.

\normalsize

\subsection{Queries With Multiple SQL OR Conditions}

We analyzed the train and dev examples from Spider looking for queries using disjoint conditions i.e., queries having  SQL "OR''  conditions. We found that \(208\) out of the \(7,000\) example queries (\(2\%\)) had conditions with only one SQL "OR''  condition, i.e., the "WHERE''  clause had two conditions connected by an SQL "OR''  condition.

In the examples queries from the IN DB, presented in \cref{sec:dbs}, we found that several concepts were mapped to more than two disjoint conditions. For instance, the concept of \emph{Outstanding} in a question translates to a query where the "invoice.status''  was equal to any of five values ( "R'' or "P'' or "A'' or "N'' or "D''), as well as having "invoice.created\_new'' equal to "Y''. \Cref{fig:queryOR} shows an example query of this type from the IN DB.

\subsection{Queries With CTE}
CTE, a.k.a. subquery refactoring, is an SQL mechanism where a "WITH''  clause defines a temporary data set whose output is available to be referenced in subsequent queries. 
We did not find these types of SQL queries in the train or dev splits from the Spider dataset. Then, to investigate how these queries will be handled by the selected models, we created a small dataset of question and query pairs using Common Table Expressions (CTE). \Cref{fig:with_query} shows an example of these types of queries.

\begin{table}[t!]
\centering
\begin{tabular}{||c | c | c | c||}
\hline
Model &  DB & EM (\%) & EX (\%) \\ 
\hline\hline
T5-Large & Spider &  64.3 & 68.7 \\ 
\hline
\end{tabular}
\caption{Training T5-Large to the Spider dataset (T5-LS).}
\label{tab:t5ls}
\end{table}

\begin{table}[ht!]
\centering
\begin{tabularx}{\columnwidth}{||
>{\centering\arraybackslash}X | 
>{\centering\arraybackslash}X |
>{\centering\arraybackslash}X |
>{\centering\arraybackslash}X |
>{\centering\arraybackslash}X |
>{\centering\arraybackslash}X||}
\hline
Model &  DB & EM (\%) & EX (\%) & EM (\%) & EX (\%) \\ 
\hline\hline
\multirow{3}{0.5in}{T5-LS} & HR & 7.69 & 8.97 & 33.3 & 44.8 \\ 
    & WH &  2.5 & 5.0 & \textbf{17.5} & \textbf{20.0} \\ 
    & IN &  -- & 6.5 & -- & 13.0 \\
\hline
\multirow{3}{0.5in}{Picard-PG} & HR & 8.9 & 17.9 & \textbf{35.9} & \textbf{60.3} \\ 
    & WH & 5.0 & 5.0  & 17.5 & 20.0 \\
    & IN &  -- & 8.7 & -- &  \textbf{15.2} \\
\hline
\end{tabularx}
\caption{Zero-Shot Learning with PD and T5-Large fine-tuned on Spider (T5-LS).}
\label{tab:zeroshot}
\end{table}

% \vspace{-0.5in}
 
\section{Experiments}
\label{sec:exp}

\subsection{Methodolody}
\label{subsec:method}

We evaluated the T5 and Picard models in two scenarios: (1) zero-shot setting, where the pre-trained model on Spider dataset was used only for testing, and (2) domain adaptation setting, where the fine-tuned model on Spider was used to continue the training on data for each of the independent DBs described in \cref{sec:dbs}. Notice that our goal is not to achieve state-of-the-art performance in the text-to-SQL task, but rather to study the  model’s performance for the different SQL queries.

\subsubsection{Model Configuration}
\label{subsec:config}

We fine-tuned the T5 Large model on the Spider dataset (T5-LS) and took the best performing model for testing (see \Cref{tab:t5ls}). For the zero-shot learning, we evaluated the pre-trained Spider model on the test set for each DB; For the domain adaptation setting, we took the pre-trained Spider model and continued its training on the corresponding training data for each schema. Each model was fine-tuned for \(512\) epochs. In our experiments we saw that training the model for more  than \(448\) epochs did not improve the model performance.

We also considered two model settings: with Disabled DB (D-DB) access, where the model did not have access to the DB content; and with Enabled DB (E-DB) access. We implemented and used a new mechanism to extract the DB values from an input question and describe it in detail in \Cref{sec:values}.

For training, we used Adafactor \citep{Shazeer2018:Adafactor}, a learning rate of \(10^{-5}\), and a batch size per device of \(5\). During testing, we enabled Picard with the highest parsing mode.

\begin{center}
\begin{table}[t!]
\small
\centering
\begin{tabularx}{\columnwidth}{||l|r|r|r|r|r||} 
\hline
\multicolumn{2}{c}{} & \multicolumn{2}{c}{Dis-DB} & \multicolumn{2}{c}{Ena-DB}\\
\hline
 Model &  DB & EM(\%) & EX(\%) & EM(\%) & EX(\%) \\
 \hline
\multirow{3}{0.3in}{T5-LS}
    & HR & 57.7  & 67.9 & 58.9 & 70.5 \\ 
    & WH & 55.0  & 60.0 & 55.0 & 62.5 \\ 
    & IN & -- & \textbf{56.5}  & -- &  52.2 \\ 
\hline
\multirow{3}{0.3in}{PD-PG}    & HR &  \textbf{62.8} & 67.9 & 62.8 & \textbf{71.8}  \\ 
    & WH &  60.0 & 67.5 & 62.5 & \textbf{70.0} \\
    & IN &  --  & 26.1 & -- & 26.1\\
\hline
\end{tabularx}
\caption{Domain Adaptation of T5-LS and PD to selected DB.}
\label{tab:dares}
\end{table}
\end{center}

\subsection{Results}
In this section we present the results of our investigation of the T5 and Picard models with the DBs presented in \Cref{sec:dbs}.

\subsubsection{Zero-Shot Evaluation}
\Cref{tab:zeroshot} presents the evaluation results for the T5-large model fine-tuned to the Spider data (T5-LS), and for the Picard model, using T5-LS as a base model, and configured with the highest parsing mode in Picard - "parse\_with\_guards''  (PG). 

From this experiment we see that for the HR and IN DBs the Picard model improves over the T5-LS model, even though none of the models have domain adaptation on these DBs. However, we did not see an improvement for the WH DB when using Picard compared to the base T5-LS model.
For the IN DB queries, we could not run the EM because the SQL parser threw errors when evaluating some of the queries from this DB having multiple SQL conditions enclosed in parentheses. We also see that the models in a zero-shot learning perform better on simpler databases such as HR, but the performance drops for DBs having a more complex schema involving multiple tables, such as WH and IN. We show some examples of challenging queries in \Cref{sec:zerochall}.

\subsubsection{Domain Adaptation for Independent Schemas}

We investigated how domain adaptation of the T5 and Picard models perform for the selected DBs. We evaluated the T5-LS model fine-tuned to the HR, WH, and IN DBs for two settings: without and with Picard.  For Picard we used the highest parsing level "parse\_with\_guards'' (PG). We also included the results of the model when the DB content was disabled (Dis-DB) or enabled (Ena-DB). The results shown in \Cref{tab:dares} indicate the advantage of using Picard over the T5-LS model for the HR and WH DBs. However, for the IN DB we see a decrease in the performance when Picard is used compared to the base T5-LS model. We checked the predictions from the Picard model for the IN DB and noticed that many were incomplete or empty. We think this is because the parsing algorithm in Picard does not currently handle queries with multiple conditions including parenthesis, such as the one in \cref{fig:queryOR}, common in queries from the IN DB.

\begin{center}
\begin{table}[t!]
\small
\centering
\begin{tabularx}{\columnwidth}{l|r|r|r|r|r} 
\hline
& \multicolumn{4}{c}{Model}\\
 & T5-LS & P-PG & P-PwoG & P-Lex\\
\hline
  DB  & EX (\%)  & EX (\%) & EX (\%) & EX (\%) \\
\hline
    HR-Fnc    & \textbf{62.8}  & 17.9 & 17.9 & 11.5 \\ 
    HR-WITH   & \textbf{50.0}  & 0.0  & 0.0  &  0.0 \\
\hline
    WH-Fnc    & \textbf{57.5}  & 5.0 & 5.0 & 5.0   \\
    WH-WITH   & \textbf{42.9}  & 9.1 & 9.1 &  9.1  \\

\hline
\end{tabularx}
\caption{Domain Adaptation of T5-LS and Picard to special queries using SQL String functions or CTE from selected DBs.}
\label{tab:daspecial}
\end{table}
\end{center}

% \vspace{-0.3in}
\subsubsection{Performance on Special SQL Queries}

In this section we focus our attention on the performance of the models with the special queries presented in \Cref{sec:queries}. We ran the experiments enabling the access to the DB content since in our previous experiments we saw better performance for this configuration, and included the three parsing levels in Picard:  parsing with guards (P-PG), parsing without guards (P-PwoG), and lexing (P-Lex).

\paragraph{Queries with String Functions.}
To test how well the models could learn SQL functions in queries, we created a modified version of the HR and WH datasets: HR-Fnc and WH-Fnc,  respectively. The modification was done in the queries to introduce the \textit{trim} and \textit{lower} SQL functions, without modifying the original questions. The number of examples in each split for the HR-Fnc and WH-Fnc versions of the dataset were the same as in the original HR and WH dataset described in \Cref{tab:splits}. We show the results for the T5-LS model, and for the Picard model using each of the three Picard parsing levels (Picard-PG, Picard-PwoG, and Picard-Lex) in \Cref{tab:daspecial}. Here we notice that the T5-LS model gives the best performance for both HR-Fnc and WH-Fnc, compared to when Picard is available in any of its parsing levels. 

\paragraph{Queries using the WITH clause.}

Questions comparing different columns values could result into SQL queries using the "WITH'' clause, a.k.a., Common Table Expression (CTE). From real scenarios, we found questions making comparisons to be very common. To analyze how the models (T5-LS and Picard) handle this types of queries, we created a small sample of comparative questions and their SQL queries with CTE, for the HR and WH DBs. The HR-WITH corpus have examples of CTE queries with splits \(35, 4,\) and \(8\) for train, dev, and test, respectively.  For the WH-WITH the splits are \(18\), \(3\), and \(7\) for the train, dev, and test, respectively. In \Cref{tab:daspecial} we can see that for queries with CTE the Picard model has a low performance for both HR-WITH and WH-WITH corpus, compared to the base T5-LS model. Although the splits have a small amount of examples, our intention was not train the best model but to have an idea of the performance of the models for this type of queries. For Picard model, the lower performance is because it cannot parse correctly the SQL structure of the CTE queries.

\paragraph{Queries with multiple conditions.}

The examples questions and queries from the IN DB include several instances where concepts in a question need to be mapped to multiple conditions in the SQL query. %\Cref{cc} shows the frequency of different types of queries for this DB. 
\Cref{tab:dares} shows that the domain adaptation for the IN DB improved the performance for both T5 and Picard models for this schema. However, we can see that using Picard, with parsing level "parse\_with\_guards'', reduces the performance. This is mainly because the current Picard parsing does not take into account SQL conditions involving multiple conditions enclosed in parenthesis, then the predictions are not generated or are incomplete when using Picard. For this type of queries, the highest execution accuracy of \(56.5 \% \) was given by the base T5-LS model.

\subsection{Lessons Learned and Challenges Found}

In this section we share the lessons we learned while doing domain adaptation for our three DBs.

\subsubsection{Lessons}
\paragraph{Lesson 1. SQL Query Structures.}
\label{sec:lcqueries}
We were interested in questions that generated SQL queries having SQL functions such as string functions (e.g., lower, trim), \(WITH\) clauses, and conditions enclosed in parenthesis to override operator precedence rules. 

We found that queries using these SQL structures, not included in Spider examples, still need to be supported. Specifically, we noticed that while Picard delivers good performance for simple and complex questions from simple (HR) and complex DBs (WH), but when given queries with SQL structures not previously considered, the model is not ready yet to process such queries, as it considers invalid tokens from functions, \(WITH\) clauses, and parenthesis from conditions in such SQL queries.  

\paragraph{Lesson 2. Domain Adaptation with a Small Corpus.}
Our experiments show that domain adaptation for Text-to-SQL can achieve good performance results for new DBs when leveraging large text-to-SQL models such as T5-Large or Picard, pre-trained on the Spider dataset.

Our results confirm previous studies \citep{Lauscher2020:zero} that have shown Few-shot learning as an effective strategy for the Text-to-SQL task for SQL queries using similar language elements than those in the Spider dataset. However, we also saw that for SQL queries with different structures than the ones considered in the Spider dataset (e.g., with string functions, using WITH clause, or multiple compound conditions) the performance decreases significantly when using a couple tens of example queries from those type of structures. This is a sample that more diversity of SQL query structures could be incorporated into larger datasets used for training and testing of Text-2-SQL models.

\paragraph{Lessons 3. Continue Training on Pre-trained Text-to-SQL Large Model.}
When fine-tuning a model for a new DB, we had two options using the selected base model: (1) Fine-tune the model on the Spider dataset and the new DB altogether, or (2) Fine-tune the the model on the Spider dataset first, select the best model, and then continue the training on the new DB data. 

We achieved similar performance with these two fine-tuning options. However, we opted for the second option, where we continued the training of the pre-trained Spider model for the specific DB schema using a couple tens or hundreds of examples. By selecting this option we reduced significantly the training time for a new DB from a couple of hours to a couple of minutes.

\subsubsection{Challenges}

\paragraph{Challenging Knowledge Grounding Without Domain Adaptation.}

Although large pre-trained models can provide a good baseline for the text-to-SQL task, without domain adaptation some examples queries could become difficult, even impossible, to predict correctly when trying to infer queries for questions of a not previously seen DB.

\paragraph{Logic Grounding.}

Sometimes models need to learn the correct mapping between a concept and a set of SQL conditions, we called this \textit{Logic} grounding. In \Cref{sec:zerochall}, example (2) in \Cref{tab:zero-examples} shows an example of this logic grounding. In the example, the gold query maps the concept of "outstanding" to several SQL conditions, but the model fails to capture the conditions associated with this concept from the IN DB in the Zero-shot setting. Doing domain adaptation with a couple tens or hundreds of examples does help the model to learn this mapping between the concept and the SQL condition.

\paragraph{Diversity of SQL structures in Datasets.} Large-scale annotated datasets are a great resource to fine-tuning models on the text-to-SQL task. However, more efforts are still needed to increase the richness and diversity of the example queries in these datasets to include different SQL supported elements. 

\paragraph{Evaluation Metrics.}
Evaluation metrics based on SQL components still need to be improved to consider different SQL queries structures.
Exact Match (EM), Execution Accuracy (EA), and Test Suite Accuracy (TSA) are the most popular metrics used to measure the performance of Text-2-SQL models. However, when we tested queries having a format including nested parentheses on the condition, using SQL functions for string manipulation, or special clauses such as WITH, the current implementation of the EM metric for the Spider dataset threw errors. We analyzed the errors and found that the SQL parser does not take into account these SQL structures, so it fails to correctly process the SQL even if the query is correct.
\citeauthor{Hazoom2021:WildStack} introduced the Partial Component Match F1 (PCM-F1) metric using a SQL parser (JSqlParser) covering more variations of SQL queries \citep{Hazoom2021:WildStack}. However, the PCM-F1 metric is still prone to false negatives and the underlying SQL parser could be improved further.

\section{Conclusions}

Deep Learning Models approaches for the Text-to-SQL task have been continuously emerging in the last years. We have empirically investigated the performance of one of the top models on the large-scale Spider English dataset in two scenarios, Zero-shot and Domain Adaptation. We shared the lessons learned while doing domain adaptation for three independent DBs not included in the Spider dataset splits. We found some SQL query structures for which the top performing model has not been adapted to and highlight SQL parsing issues for the Picard model and for the exact match metric. We plan to address these shortcomings in order to increase the coverage and performance of the model.

\bibliography{custom}
\bibliographystyle{acl_natbib}

% This is an appendix.
\appendix

\section{Mapping Values in Questions to Database (DB) Columns}
\label{sec:values}

\begin{center}
\begin{table*}[t!]
\centering
\begin{tabular}{l|l}
TRF & LRF  \\
\hline
root=\small{$ prop\_owner\_VAR1\_has\_VAR2\_ $}& 
root=\small{$prop\_owner\_ product\_has\_price\_$} \\
\small{-> has  [ hasPartOfSpeech("verb"), hasLemmaForm("have")]}
 & \small{-> has   [hasPartOfSpeech("verb"), hasLemmaForm("have")]}\\
 \small{ subj -> VAR1 [!hasParseFeature("ving")},
 & \small {subj -> product [!hasParseFeature("ving"),} \\
 \small{hasLemmaForm("VAR1") ] }
 &  \small{hasLemmaForm(“product") ] } \\
 \small { obj -> VAR2  [ hasLemmaForm("VAR2")]}
& \small { obj -> price[ hasLemmaForm(“price")]}
\\
\hline
\end{tabular}
\caption{Example of TRF and LRF during the adaptation process in the NLP engine.}
\label{tab:trf_example}
\end{table*}
\end{center}
The PD model allows to use or ignore the DB content during the question serialization process. When the DB content is enabled, values in the input question are mapped to columns in the DB through a value search algorithm. Overall, the model performs better when the values in the question are identified and attached along with the corresponding table columns in a given DB. However, the default mechanism in PD used to identify values from an input query requires scanning question tokens and all tables' columns values from a DB in an online fashion. This mechanism works well for small DBs but it is not ideal for a real scenario where the time required to find values can grow linearly with the size of the DB. 

Instead of scanning every token in the input question for matching values in a DB, we leverage an intermediate representation of the semantic concepts in a DB through an ontology-based mechanism, to identify the schema linking between values and table columns from an input question. The NLP system to extract the column-value mapping is described in \Cref{sec:appendix}.
% along with a dictionary mapping of the DB content

The advantage of the ontology-based column-value mapping mechanism is that we do not need to query the DB multiple times searching for the table-column pairs for a given value in a question. Instead, we run the NLP engine to obtain the intermediate representation with populated values, extract the column-value mapping from each table found for the given question, and serialize this information along with the input question as a tagged sequence using the BRIDGE \citep{Lin2020:Bridge} mechanism.

\subsection{Value Disambiguation Engine}
\label{sec:appendix}

The value disambiguation engine is an NLP engine that receives as an input a question and a DB and returns a mapping between values in the question and the columns in the DB where values belong to.
\begin{center}
\begin{table*}[h!]
\centering
\begin{tabularx}{\textwidth}{|| l|l|l || }
\hline
  Question & Gold SQL & Predicted SQL\\
\hline
\small
1. "What projects are open?"
  & \small SELECT DISTINCT p\_id & \small select distinct p\_date, t1.cu\_name, t2.cu\_addr \\
  & \small FROM \textbf{p\_number} WHERE \textbf{status=`A'} & \small from \textbf{inv} as t1 join contract as t2 \\
  &  & \small on t1.cus\_address = t2.cus\_address \\
  & & \small where \textbf{t2.con\_status = `Open'}\\
\hline
 \small 2. "Show outstanding invoices" & 
 \small SELECT id, bill\_amnt FROM inv & \small select count(*) from inv as t1 \\
 & \small \textbf{WHERE} (status=`RT' OR status=`RJ' & \small join contract as t2 \\
 & \small OR status=`P' OR status=`A') & \small on t1.con\_number = t2.con\_number \\
 & \small AND created=`Y' & \\
\hline
\end{tabularx}
\caption{ Examples of Challenging Knowledge Grounding for Models in a Zero-Shot Learning Scenario.}
\label{tab:zero-examples}
\end{table*}
\end{center}
\subsection{Data Items and Column-Value Mapping}
To map columns to values in a natural language question we rely on data-items. An NLP engine takes an input question and generates a set of structured statements, called data-items. Data-items are an intermediate representation of the semantic concepts in the input English question or sentence. To create data-items the NLP engine uses an adaptation process and query processing module.

\paragraph{Adaptation.} The adaptation phase adapts the system for a specific schema. 
This phase requires two resources: a template-rules file (TRF), and a schema-annotation file (SAF). The TRF is a text file containing sub-tree patterns representing English syntax rules. This file is domain-independent and doesn’t contain any domain specific lexical information. The SAF is a text file that describes the domain-specific schema, it is the ontology of the target DB were the relationships that exist between the columns in each table are described. Each entry in the SAF contains a simple English phrase describing the semantic relation between two entities in the database schema, and the properties of the two entities, e.g. table-name, column-name, data-type. For instance, for a Warehouse schema a SAF entry looks like the following:

\small{
\begin{verbatim}
product has price; tableName1 is PRODUCTS;
colName1 is PRODUCT_ID; dataType1 is 
integer; tableName2 is PRODUCTS; colName2 
is PRICE; dataType2 is decimal;
\end{verbatim}
}

\normalsize
During the adaptation process the TRF rules are applied to the SAF entries and produce a lexical rule file (LRF). LRF has the same format of the TRF but includes the schema’s lexical information. Table \ref{tab:trf_example} shows an example of the TRF and corresponding LRF for the SAF entry shown above. For each SAF entry, which describes a basic sematic relation between two entities, multiple number of rules, representing a wide range of syntaxes, are automatically created.
\paragraph{Query Processing.} During the query processing phase the LRF is loaded into the memory and a subtree-pattern-matching algorithm is used to find the rules that match the input query syntax. These rules are used to detect the different entities and their semantic relations of the input natural language query. These entities are coded in a semi-structured list of items in a JSON string. The key for each item is its table-name and column-name, the value contains the properties of the column plus the syntactical position of the entity in the sentence. The semi-structured list is then processed into structured data-items. For instance, for the question \textit{"How many products have price higher than $100$?"} of a Warehouse schema, the generated data-items are as follow:

\small
\begin{verbatim}
[PRODUCTS].[PRODUCT_ID]={aggrFlag=1, 
dataType=integer, focus=select, 
aggrFunction=countDistinct}
[PRODUCTS].[PRICE]={ filterFlag=1,
value=100, dataType=decimal,
operator=greaterThan}
\end{verbatim}

\normalsize
\paragraph{Column-Value mapping.} From the example above, we can see that data-items capture values, and the column they belong to, from an input query. The NLP engine uses a pre-populated dictionary with the content from the DB, checks the dictionary, finds matching values in the sentence, and adds a filter for them in the data-items. The advantage of using data-items is that instead of querying the DB multiple times searching for the table-column pairs for a given value in a question, we run the NLP engine to obtain the data-items, and extract from them the values from each table column that are present in a given question.

\section{Challenging Knowledge Grounding in a Zero-Shot Learning}
\label{sec:zerochall}

\Cref{tab:zero-examples} presents as example two instances of  queries from our independent IN DB that show how challeging is for the models to predict the correct queries for new DBs in a Zero-Shot learning setting. In the first example, the concept of \textit{open} in the golden query is associated with the status of a project being equal to \emph{'A'}, while the predicted query associates the status to the value from in the question. In the second example, the concept of outstanding requires the query to check different statuses of the invoice, while the predicted query do not consider any condition for this concept.

\end{document}